\documentclass{IET-Conf-Paper}

\makeatletter
\long\def\@tablecaption#1#2{%
  \par
  \setbox\tablecaptionbox=\vbox{%
    \hsize=\linewidth
    \noindent #1\enskip #2\par
  }%
  \unvbox\tablecaptionbox\par
  \vspace{4pt}%
}
\makeatother

\usepackage{amsmath}
\usepackage{booktabs}
\usepackage{multirow}
\usepackage[nolist]{acronym}

\usepackage{url}

\DeclareMathOperator*{\argmin}{arg\,min}

\begin{document}
\acrodef{ml}[ML]{machine learning}
\acrodef{emt}[EMT]{electromagnetic transient}
\acrodef{cv}[CV]{cross-validation}
\acrodef{auc}[AUC]{area under the curve}
\acrodef{roc}[ROC]{receiver operating characteristic}
\acrodef{aucroc}[AUC--ROC]{area under the receiver operating characteristic curve}
\acrodef{ci}[CI]{confidence interval}
\acrodef{or}[OR]{odds ratio}

\acrodef{gbt}[GBT]{gradient-boosted tree}
\acrodef{cnn}[CNN]{convolutional neural network}
\acrodef{logreg}[LogReg]{logistic regression}
\acrodef{seqoc}[SeqOC]{sequence-current starter}

\acrodef{sir}[SIR]{source-to-line impedance ratio}
\acrodef{ftr}[FTR]{false-trip rate}
\acrodef{mtr}[MTR]{missed-trip rate}

\acrodef{slg}[SLG]{single-line-to-ground}
\acrodef{ll}[LL]{line-to-line}
\acrodef{llg}[LLG]{double-line-to-ground}
\acrodef{lll}[LLL]{three-phase}

\acrodef{dft}[DFT]{discrete Fourier transform}
\acrodef{rms}[RMS]{root-mean-square}
\acrodef{cpu}[CPU]{central processing unit}
\acrodef{gpu}[GPU]{graphics processing unit}
\acrodef{hv}[HV]{high-voltage}

\acrodef{dfg}[DFG]{Deutsche Forschungsgemeinschaft}

\title{Protection-Oriented Assessment of Machine Learning-Based Fault Detection in Power Systems}

\author{Julian Oelhaf\ad{1},
        Georg Kordowich\ad{2},
        Changhun Kim\ad{1},
        Christian Bergler\ad{3},
        Andreas Maier\ad{1},
        Johann J\"{a}ger\ad{2},
        Siming Bayer\ad{1}}

\address{%
  \add{1}{Pattern Recognition Lab, Friedrich-Alexander-Universit\"{a}t Erlangen-N\"{u}rnberg, Erlangen, Germany}
  \add{2}{Institute of Electrical Energy Systems, Friedrich-Alexander-Universit\"{a}t Erlangen-N\"{u}rnberg, Erlangen, Germany}
  \add{3}{Department of Electrical Engineering, Media and Computer Science, Ostbayerische Technische Hochschule Amberg-Weiden, Amberg, Germany}
  }

\keywords{POWER SYSTEM PROTECTION, FAULT DETECTION, MACHINE LEARNING, SECURITY, DEPENDABILITY}

\begin{abstract}
Machine learning (ML)-based fault detectors are increasingly proposed for power system protection, yet evaluations commonly rely on accuracy, F1 score, or area under the receiver operating characteristic curve (AUC--ROC) without reflecting the different protection consequences of false and missed trips. This paper develops a protection-oriented detector assessment and applies it to three ML detectors and a sequence-current starter reference on the PROTECT-90 electromagnetic transient dataset comprising 9,022 simulated fault episodes on a 90\,kV double-circuit line. The assessment combines (i) a five-condition window taxonomy with causal window-end labelling, (ii) episode-grouped outer cross-validation with an inner validation split, (iii) protection-oriented metrics for security, dependability, detection timing, decision margin, and episode-level nuisance assertions, and (iv) constrained operating-point selection under a fixed security budget. A gradient-boosted tree on engineered features, a compact one-dimensional convolutional neural network on raw waveforms, and logistic regression all achieve AUC--ROC values above 0.999, yet at a matched security budget of $10^{-3}$ their developed-fault missed-trip rates span more than two orders of magnitude, from 0.007\,\% to 1.99\,\%. Post-clearing false-trip rates of approximately $0.12$--$0.16\,\%$ per window coexist with nuisance assertions in $11.3$--$19.1\,\%$ of episodes, highlighting the importance of episode-level assessment. Among the analysed physical descriptors, missed-trip errors are most consistently associated with higher fault resistance within the sampled range. An analytical current proxy suggests a fault-current gradient that is only partly reproduced when current is derived directly from the waveforms (rank correlation 0.62). No single topology state or faulted phase consistently dominates within the sampled operating domain. Because PROTECT-90 contains fault episodes only, security is assessed on pre-fault and post-clearing windows rather than independent non-fault disturbances.
\end{abstract}

\maketitle
\begingroup
\renewcommand{\thefootnote}{}
\footnotetext{This paper is a preprint of a paper accepted by the 21st International Conference on Developments in Power System Protection (DPSP APAC 2027) and is subject to Institution of Engineering and Technology Copyright. When the final version is published, the copy of record will be available at the IET Digital Library.}
\endgroup

\section{Introduction}

Protective relaying must balance \emph{security} (no unnecessary operations) and \emph{dependability} (no missed faults)~\cite{ieee_ieee_2026,blackburn_protective_2014,horowitz_power_2023}. \Ac{ml} fault detectors promise faster operation and improved discrimination under high-resistance or evolving faults~\cite{kezunovic_role_2013}, but evaluations commonly emphasise accuracy, precision, recall, and AUC--ROC~\cite{oelhaf_scoping_2025,porawagamage_review_2024,dash_role_2024,fawcett_introduction_2006,davis_relationship_2006}, which do not directly characterise protection-specific security, dependability, or timing and may obscure whether aggregate discrimination translates into acceptable detector behaviour.

An unnecessary trip can disconnect healthy equipment or interrupt power transfer, whereas a missed trip allows fault current to persist, risking cascading outages, equipment damage, and personal injury~\cite{north_american_electric_reliability_corporation_2025_2025,sticht_investigation_2021}. IEEE~C37.113 treats security and dependability as central transmission-line protection considerations~\cite{ieee_ieee_2026}, while NERC defines protection-system misoperation by whether the composite system operates as intended~\cite{north_american_electric_reliability_corporation_misoperation_2025}. Because we evaluate detector decisions rather than complete relay, breaker, and coordination logic, false and missed trips are detector-level indicators of potential misoperation, not formal protection-system misoperations. Timing is also intrinsic, yet many \ac{ml} studies omit detection delay and the implications of causal windowed inference.

Our recent standardised framework makes the inference setting of \ac{ml}-based protection studies explicit across seven dimensions, including timing, sample validity, validation, and evaluation outputs~\cite{oelhaf_standardized_2026}. Building on that framework, the present study develops a protection-oriented detector assessment focused specifically on security, dependability, timing, and constrained operating-point selection.

This paper makes three contributions:
\begin{enumerate}
  \item A causal five-condition window taxonomy that separates deep and near pre-fault security, fault-onset sensitivity, developed-fault dependability, and post-clearing security.
  \item An episode-grouped evaluation protocol with an inner validation split for preprocessing, calibration, and threshold selection, followed by protection-oriented assessment on held-out outer folds.
  \item A metadata-stratified error analysis that relates detector-level false- and missed-trip rates to measured and estimated fault current, source-to-line impedance ratio, fault resistance, fault type, topology, and faulted phase.
\end{enumerate}

We apply this assessment to the public PROTECT-90 dataset~\cite{oelhaf_protect-90_2026} of 9,022 simulated fault episodes on a 90\,kV double-circuit line. Our prior work addressed fault detection and line identification~\cite{oelhaf_systematic_2025}, fault classification and localisation~\cite{oelhaf_controlled_2026}, and robustness to measurement and communication degradation~\cite{oelhaf_robustness_2026}; here we focus on detector security, dependability, timing, and operating-condition dependence.

\section{Dataset and physical descriptors}
\label{sec:dataset-scope}

PROTECT-90~\cite{oelhaf_protect-90_2026} contains 9,022 one-second electromagnetic-transient fault episodes from a 90\,kV double-circuit line, sampled at 6,400\,Hz with 48 voltage and current channels at eight relay locations. Its manifest provides fault inception and clearing times, fault type, resistance, location, source strength, topology, and operating-condition parameters. Faults include single-line-to-ground (SLG), line-to-line (LL), double-line-to-ground (LLG), and three-phase (LLL) cases with $R_f\in[0,10]\,\Omega$. No independent fault-free disturbance episodes are included, so security is assessed only on pre-fault and post-clearing windows and does not cover restraint under transformer energisation or switching transients.

To relate missed-trip behaviour to electrical operating conditions, we compute the \ac{sir} following C37.113~\cite{ieee_ieee_2026}:

\begin{equation*}
\mathrm{SIR} = \frac{V_n^2/S_\mathrm{sc}} {L\sqrt{r^2+x^2}},
\end{equation*}

where $V_n=90$\,kV, $S_\mathrm{sc}$ is the dominant-source short-circuit power, $L$ is the faulted-line length, and $r$ and $x$ are its positive-sequence resistance and reactance per unit length.

Because \ac{sir} and $R_f$ describe detectability only indirectly, we also compute a simplified fault-current proxy $\hat{I}_f$. Let $V_\mathrm{ph}=V_n/\sqrt{3}$, and let $Z_s$, $Z_\ell$, and $Z_{\ell0}$ denote the source, positive-sequence line, and zero-sequence line impedances to the fault. For this magnitude-only proxy, positive- and negative-sequence line impedances are taken equal, all source-sequence impedances are approximated by the same short-circuit-power-derived magnitude, and an in-service secondary source is represented by a parallel source-impedance reduction. The proxy is

\begin{equation*}
\hat{I}_f =
\begin{cases}
\dfrac{V_\mathrm{ph}}
{\left|Z_s+Z_\ell+R_f\right|},
& \mathrm{LLL},\\[6pt]
\dfrac{\sqrt{3}V_\mathrm{ph}}
{\left|2(Z_s+Z_\ell)+R_f\right|},
& \mathrm{LL},\\[6pt]
\dfrac{3V_\mathrm{ph}}
{\left|2(Z_s+Z_\ell)+(Z_s+Z_{\ell0})+3R_f\right|},
& \text{ground fault}.
\end{cases}
\end{equation*}

The proxy is approximate, particularly for LLG faults. We therefore also derive a waveform descriptor $I_f$: the maximum root-mean-square phase current across all eight locations and three phases over the first complete 20\,ms cycle after fault inception; the preceding cycle is retained as a pre-fault reference. Across 8,618 eligible episodes, $I_f$ spans 283--27,395\,A and shows only moderate rank agreement with the proxy ($\rho=0.62$, Spearman, $p<10^{-300}$; 0.59--0.67 by fault type). We therefore use $I_f$ for operating-condition stratification.

\section{Assessment protocol}

Each episode is segmented into 20\,ms windows of 128 samples. The end of each window, $t_\mathrm{end}^w$, defines the causal decision time because the prediction uses only samples observed up to that point. Relative to the fault interval $[t_\mathrm{start},t_\mathrm{end}]$, each window is assigned to one of five conditions:
\[
\begin{array}{@{}ll@{}}
\text{deep pre-fault:} & t_\mathrm{end}^w \leq t_\mathrm{start}-40\,\mathrm{ms},\\
\text{near pre-fault:} & t_\mathrm{start}-40\,\mathrm{ms}<t_\mathrm{end}^w<t_\mathrm{start},\\
\text{fault onset:} & \text{first window with } t_\mathrm{end}^w\geq t_\mathrm{start},\\
\text{developed fault:} & \text{subsequent windows with } t_\mathrm{end}^w\leq t_\mathrm{end},\\
\text{post-clearing:} & t_\mathrm{end}^w\geq t_\mathrm{end}+20\,\mathrm{ms}.
\end{array}
\]

Fault-onset and developed-fault windows are labelled 1; all others are labelled 0. Onset is reported separately because it may contain only a partial fault signature, and windows within the 20\,ms post-clearing guard band are excluded. Episodes with fault durations shorter than one window remain in the onset and security assessments but are excluded from developed-fault denominators and timing analysis. This eligibility rule also governs operating-point selection, operating-domain maps, and fault-current stratification.

\subsection{Validation and operating point}

Evaluation uses five-fold episode-grouped outer \ac{cv}; within each outer training fold, 15\,\% of episodes form the inner validation set for calibration and threshold selection. Episodes remain indivisible across splits. Inner training uses non-overlapping 20\,ms windows, whereas inner validation and outer testing use a 2.5\,ms stride, matching selection and evaluation cadence. All preprocessing parameters, calibrators, and thresholds are determined without outer-test access.

The operating threshold is selected on the inner validation set as the most dependable point that satisfies a fixed security budget:

\begin{equation}
\tau^\star(\alpha) =
\argmin_{\tau \in \mathcal{T}}
\operatorname{MTR}(\tau)
\quad\text{subject to}\quad
\operatorname{FTR}(\tau)\leq\alpha ,
\label{eq:threshold}
\end{equation}

with $\alpha=10^{-3}$ fixed before outer-test evaluation. $\mathcal{T}$ contains all distinct calibrated inner-validation scores plus a sentinel above the maximum, giving an exact search and a feasible no-assertion endpoint. Among feasible thresholds minimising \ac{mtr}, ties are resolved by lower \ac{ftr} and then higher threshold, favouring security. \Ac{ftr} pools deep pre-fault, near pre-fault, and post-clearing windows; \ac{mtr} uses developed-fault windows, excluding onset. Only inner-validation \ac{ftr} is constrained; outer-test \ac{ftr} is reported as observed. Sensitivities for $\alpha\in\{10^{-4},10^{-5}\}$ and the equal-error point $\argmin_{\tau}\left|\operatorname{FTR}(\tau)-\operatorname{MTR}(\tau)\right|$ appear in Table~\ref{tab:app-sensitivity}. The fixed budget makes the operating point an explicit protection choice and enables matched-security comparison.

\subsection{Protection metrics}

A \emph{trip} is a positive detector indication at the selected threshold, not a confirmed breaker operation. Security is measured by \ac{ftr} over pooled deep pre-fault, near pre-fault, and post-clearing windows and separately by condition; dependability uses developed-fault \ac{mtr}, with onset reported separately. Episode-level failure-to-trip rate is the fraction of eligible episodes without a trip before a specified deadline.

Detection delay runs from fault inception to the end of the first post-inception window producing a trip and excludes measurement filtering, communication, inference runtime, breaker operation, and other implementation delays. Predictions occur every 2.5\,ms, but asynchronous fault inception and aggregation across episodes mean reported delays need not be multiples of 2.5\,ms. Delay quantiles are conditional on a post-inception assertion, possibly after clearing; episodes without one count as failures to trip. We report $d_{50}$, $d_{95}$, and failure-to-trip rate at 40\,ms. This deadline characterises detector response, not a formal slow-trip classification, which depends on protection-system timing and coordination.

Decision margin is defined as $|p-\tau|$, where $p$ is the calibrated fault probability and $\tau$ is the operating threshold. Low-margin decisions have margins below 0.10. This measures threshold proximity, not perturbation robustness or validated probabilistic confidence.

\begin{table*}[t]
\centering
\caption{Overall detector performance (mean $\pm$ sample standard deviation over three seeds and five outer folds). Error rates are reported in percent; $d_{50}$ and $d_{95}$ denote the median and 95th-percentile algorithmic detection delay, conditional on an eventual trip.}
\label{tab:performance}
\footnotesize
\begin{tabular*}{\textwidth}{@{\extracolsep{\fill}}lccccccc@{}}
\toprule
& \textbf{Aggregate} & \multicolumn{2}{c}{\textbf{Security}} & \multicolumn{2}{c}{\textbf{Dependability}} & \multicolumn{2}{c}{\textbf{Timing}} \\
\textbf{Model} & \textbf{\acs{aucroc}} & \textbf{Pooled \acs{ftr} (\%)} & \textbf{Post-clear. \acs{ftr} (\%)} & \textbf{Developed \acs{mtr} (\%)} & \textbf{Onset \acs{mtr} (\%)} & \textbf{$d_{50}$ (ms)} & \textbf{$d_{95}$ (ms)} \\
\midrule
\acs{gbt} & $0.99999\pm0.00001$ & $0.079\pm0.028$ & $0.138\pm0.045$ & $0.007\pm0.011$ & $2.11\pm0.81$ & $1.31\pm0.03$ & $2.43\pm0.02$ \\
\acs{cnn} & $0.99970\pm0.00011$ & $0.078\pm0.051$ & $0.122\pm0.064$ & $0.008\pm0.007$ & $10.83\pm1.98$ & $1.49\pm0.04$ & $3.00\pm0.36$ \\
\acs{logreg} & $0.99931\pm0.00007$ & $0.092\pm0.022$ & $0.164\pm0.035$ & $1.99\pm0.55$ & $29.32\pm1.92$ & $1.84\pm0.05$ & $14.03\pm1.10$ \\
\acs{seqoc} & $0.96966\pm0.00138$ & $0.089\pm0.013$ & $0.161\pm0.023$ & $38.29\pm2.50$ & $99.97\pm0.04$ & $10.19\pm0.58$ & $20.68\pm0.58$ \\
\bottomrule
\end{tabular*}
\end{table*}

\section{Detectors}

We evaluate one sequence-current reference and three trained \ac{ml} models, with hyperparameters fixed before evaluation. Within each outer fold, the \ac{ml} models use inner-training data, while calibration and threshold selection use only inner validation. Calibration considers the identity map, Platt scaling, and isotonic regression~\cite{zadrozny_transforming_2002,guo_calibration_2017}; Platt scaling and isotonic regression are fitted there, and all options are compared by Brier score~\cite{brier_verification_1950}. Isotonic regression is selected in all 60 detector--seed--fold combinations. Selection uses in-sample Brier scores and does not establish out-of-sample calibration quality. The validation-selected calibrator and threshold~\eqref{eq:threshold} are applied unchanged to outer-test episodes.

The \ac{seqoc} is stateless. A single-cycle \ac{dft} produces the phase-current phasors at each measurement location. With $a=e^{j2\pi/3}$, the zero- and negative-sequence components are
\[
\begin{aligned}
I_0 &= \frac{1}{3}\left(I_a+I_b+I_c\right),\\
I_2 &= \frac{1}{3}\left(I_a+a^2I_b+aI_c\right).
\end{aligned}
\]
Its score is the maximum of $\max(|I_0|,|I_2|)$ across measurement locations. The starter therefore responds to current unbalance but has no sustained pickup quantity for an ideal balanced three-phase fault. For protocol comparability, its continuous score is calibrated and thresholded using the same inner-validation procedure as the \ac{ml} scores. It is therefore a sequence-current reference under a common evaluation protocol, not a conventionally coordinated relay setting.

The \ac{gbt} uses 304 standardised engineered features. Each of 48 waveform channels contributes root-mean-square value, absolute peak, crest factor, standard deviation, maximum absolute sample-to-sample change, and spectral-energy fraction at 400\,Hz and above (288 features). At each of the eight locations, root-mean-square values of the instantaneous three-phase current and voltage sums provide 16 additional features. Standardisation is fitted on inner-training data only. The histogram-based gradient-boosting classifier~\cite{pedregosa_scikit-learn_2011} uses learning rate 0.1, at most 300 iterations, and at most 31 leaves per tree.

The \ac{cnn} uses the standardised $(128\times48)$ waveform window and three one-dimensional convolutional blocks with 64/128/128 channels and kernels 7/7/3. Each convolution is followed by batch normalisation and a rectified linear unit; the first two blocks also use max-pooling. Adaptive average pooling precedes dropout (0.1) and a linear output layer. Training uses class-weighted binary cross-entropy and Adam ($10^{-3}$ learning rate, $10^{-4}$ weight decay, batch size 256, 12 epochs).

The \ac{logreg} uses an L2-regularised linear model with $C=1.0$ on the same standardised 304-dimensional feature vector as the \ac{gbt}. It provides a linear reference using the same engineered feature representation as the \ac{gbt}.

The \ac{cnn} is implemented in PyTorch~2.8, while the remaining detectors use scikit-learn~1.8.0. Complete configurations and library versions are stored with each run and will be released with the source code.

\section{Results}
\label{sec:results}

The evaluation comprises all 9,022 episodes in five-fold episode-grouped outer \ac{cv}, repeated over three seeds. Values are the mean $\pm$ sample standard deviation over the 15 seed--fold evaluations, with each evaluation weighted equally. Across these evaluations, each detector produces 10.4 million outer-test window predictions; adjacent predictions are strongly overlapping and are not treated as statistically independent observations.

Table~\ref{tab:performance} compares aggregate discrimination with protection-oriented security, dependability, and timing. All three \ac{ml} detectors achieve AUC--ROC values above 0.999. The validation-selected operating points yield mean outer-test pooled \ac{ftr} values close to the $10^{-3}$ budget, so mean security is approximately matched across the comparison. Under that constraint, the developed-fault \ac{mtr} differs by more than two orders of magnitude: $0.007\,\%$ for \ac{gbt} and $0.008\,\%$ for \ac{cnn}, against $1.99\,\%$ for \ac{logreg} --- roughly $280\times$ between \ac{gbt} and the linear reference and $5{,}000\times$ relative to \ac{seqoc} ($38.29\,\%$); ratios are computed from unrounded means. A separation of only $0.0007$ in AUC--ROC therefore accompanies a roughly $280$-fold difference in developed-fault missed-trip rate.

The separation between detectors increases at fault onset. \ac{gbt} misses $2.1\,\%$ of onset windows, compared with $10.8\,\%$ for \ac{cnn} and $29.3\,\%$ for \ac{logreg}. The two nonlinear detectors have similar developed-fault missed-trip rates but differ approximately fivefold at onset, showing that developed-fault performance alone does not characterise onset sensitivity. \Ac{seqoc} misses almost every onset window ($99.97\,\%$): at the threshold required by the security budget, the transient zero- and negative-sequence content present at inception does not reach pickup.

Median first-trip delays are $1.31$--$1.84$\,ms for the \ac{ml} detectors and $10.19$\,ms for \ac{seqoc}. The 95th-percentile delay is $2.43$\,ms for \ac{gbt} and $3.00$\,ms for \ac{cnn} but reaches $14.03$\,ms for \ac{logreg} and $20.68$\,ms for \ac{seqoc}. The failure-to-trip rate at 40\,ms is zero for \ac{gbt} and \ac{cnn}, $0.23\,\%$ for \ac{logreg}, and $25.5\,\%$ for \ac{seqoc}. These episode-level metrics quantify initial pickup rather than sustained dependability: \ac{seqoc} still produces an early trip on the unbalanced faults it does detect while missing $38.3\,\%$ of developed-fault windows overall. The reported values exclude implementation-level and breaker delays. For the \ac{ml} detectors, median delays are shorter than the 2.5\,ms decision interval, which is possible because fault inception is asynchronous with the decision times. The $d_{95}$ values and failure-to-trip rates distinguish the detectors more clearly.

Sustained dependability separates the detectors further. Among eligible episodes with a post-inception assertion, \ac{gbt} and \ac{cnn} subsequently de-assert while the fault is still active in $0.31\,\%$ and $0.17\,\%$ of episodes, against $11.1\,\%$ for \ac{logreg} and $16.8\,\%$ for \ac{seqoc}. For eligible episodes with developed-fault decisions, longest miss runs equal consecutive miss counts times the decision interval, including zeros. Mean within-evaluation 95th percentiles are $0$\,ms for \ac{gbt} and \ac{cnn}, $11.8$\,ms for \ac{logreg}, and $393.7$\,ms for \ac{seqoc}. Mean within-evaluation maxima are $7.8$, $3.7$, $475.0$, and $577.0$\,ms, respectively. A pickup followed by a sustained de-assertion is a different failure mode from a late pickup, and only the developed-fault view exposes it. The underlying decision sequences could also be used to evaluate persistence or $N$-out-of-$M$ logic, extending the present window-level assessment.

\subsection{Operating-point and temporal behaviour}

The validation-selected thresholds yield a mean outer-test \ac{ftr} within budget for each detector, although $17$ of $60$ detector--seed--fold combinations exceed it, at most by a factor of $1.9$. Such overshoots can arise when transferring thresholds to held-out episodes and are reported rather than corrected, because correction would require outer-test data.

False trips occur mainly after fault clearing rather than before fault inception. For \ac{gbt}, the post-clearing \ac{ftr} is $0.138\,\%$, against $0.006\,\%$ and $0.011\,\%$ in the deep and near pre-fault conditions. The same ordering holds for every detector: post-clearing rates of $0.12$--$0.16\,\%$ exceed the corresponding pre-fault rates by factors ranging from five for \ac{cnn} to more than forty for \ac{logreg}, while \ac{seqoc} produces no pre-fault false trips. The complete condition-wise security breakdown is reported in Table~\ref{tab:app-security}.

Decision margins further distinguish the onset behaviour. At fault onset, $47.6\,\%$ of \ac{logreg} decisions lie within 0.10 of its selected threshold, consistent with a detector operating close to its decision boundary at $\tau^\star=0.87$ and with weak separation when only a partial fault signature is available. The corresponding rates are $1.9\,\%$ for \ac{gbt} and $3.9\,\%$ for \ac{cnn}, indicating that their onset decisions are rarely close to the selected operating thresholds. These aggregate margin statistics do not establish whether the missed onset decisions could be recovered by a small threshold adjustment. The \ac{gbt} probability and margin distributions are shown in Fig.~\ref{fig:app-margin}.

\subsection{Episode-level nuisance assertions}

Window-level rates do not give the probability that an \emph{episode} contains a spurious assertion: with several hundred overlapping windows per episode, a small window rate can yield a substantial episode rate. We therefore report the fraction of episodes with at least one positive decision separately before the fault and after clearing.

Pre-fault nuisance assertions are rare: $0.011\,\%$ of episodes for \ac{gbt}, $0.104\,\%$ for \ac{cnn}, $0.004\,\%$ for \ac{logreg}, and none for \ac{seqoc}. Post-clearing behaviour is markedly different. At the $10^{-3}$ operating point, $11.3\,\%$ of episodes contain at least one post-clearing assertion for \ac{gbt}, $12.7\,\%$ for \ac{cnn}, $16.8\,\%$ for \ac{logreg}, and $19.1\,\%$ for \ac{seqoc}: roughly one episode in eight to one in five, despite post-clearing window false-trip rates of $0.12$--$0.16\,\%$. The post-clearing episode rate also depends on the selected operating point: at the fixed security budget, it is $2.4$--$2.9$ times that at the equal-error point for \ac{gbt} and \ac{cnn} (Table~\ref{tab:app-sensitivity}). The concentration of false assertions after clearing is consistent with residual transients. The distinction matters for interpretation: a small window-level false-trip rate can coexist with a much larger episode-level nuisance-assertion probability, which would be concealed if only the former were reported.

\subsection{Operating-condition dependence}
\label{sec:operating-condition-dependence}

\begin{figure}[ht]
    \centering
    \includegraphics[width=0.8\columnwidth]{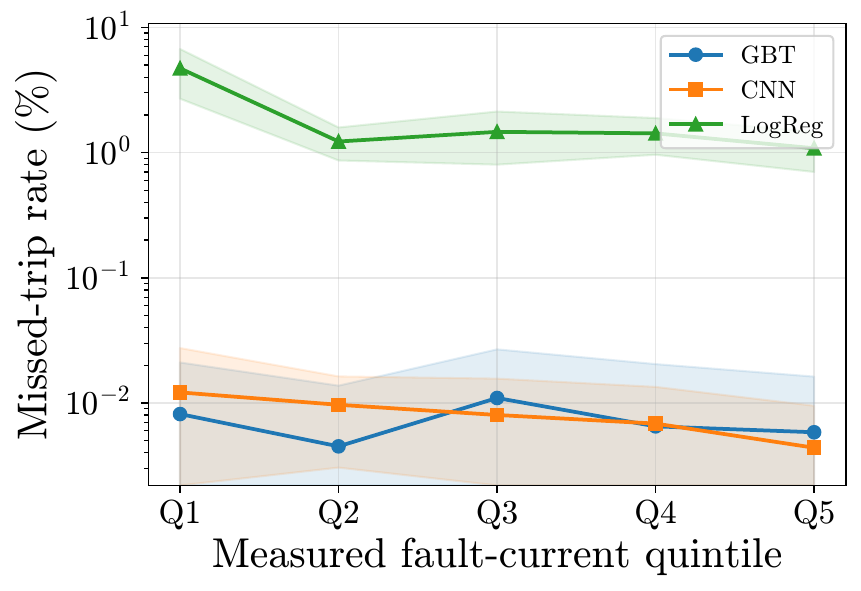}
    \caption{Missed-trip rate on developed-fault windows by measured fault-current quintile for the three \ac{ml} detectors, on a logarithmic axis: the per-quintile means span about three decades, so a linear axis would place \acs{gbt} and \acs{cnn} on the zero line. Quintiles run from the weakest to the strongest measured fault current. Lines show the mean and shaded regions $\pm$ one sample standard deviation across the 15 seed--fold evaluations, with the lower edge clipped at the axis floor: for \acs{gbt} and \acs{cnn} the sample standard deviation exceeds the mean, so the interval extends below zero.}
    \label{fig:fault-current}
\end{figure}

Figure~\ref{fig:fault-current} stratifies the developed-fault \ac{mtr} by measured fault-current quintile. The monotone decrease that the analytical proxy suggests is reproduced by only one detector: for \ac{cnn} the rate falls consistently from $0.012\,\%$ in the weakest-current quintile to $0.004\,\%$ in the strongest. For \ac{logreg} the weakest quintile is clearly worst ($4.71\,\%$ against $1.08\,\%$ in the strongest), but the intermediate quintiles are flat rather than ordered, and for \ac{gbt} the trend is absent at this resolution ($0.008\,\%$ against $0.006\,\%$, with the maximum occurring in the middle quintile). Because the analytical proxy correlates with the measured current at only $\rho=0.62$, the smooth gradient obtained with $\hat{I}_f$ is not fully reproduced when current magnitude is derived directly from the waveforms. Fault-current magnitude remains a correlate of missed trips for the weaker detectors, but it is not, on its own, a reliable predictor of \ac{ml} detector error across the sampled domain.

\begin{figure}[ht]
    \centering
    \includegraphics[width=0.8\columnwidth]{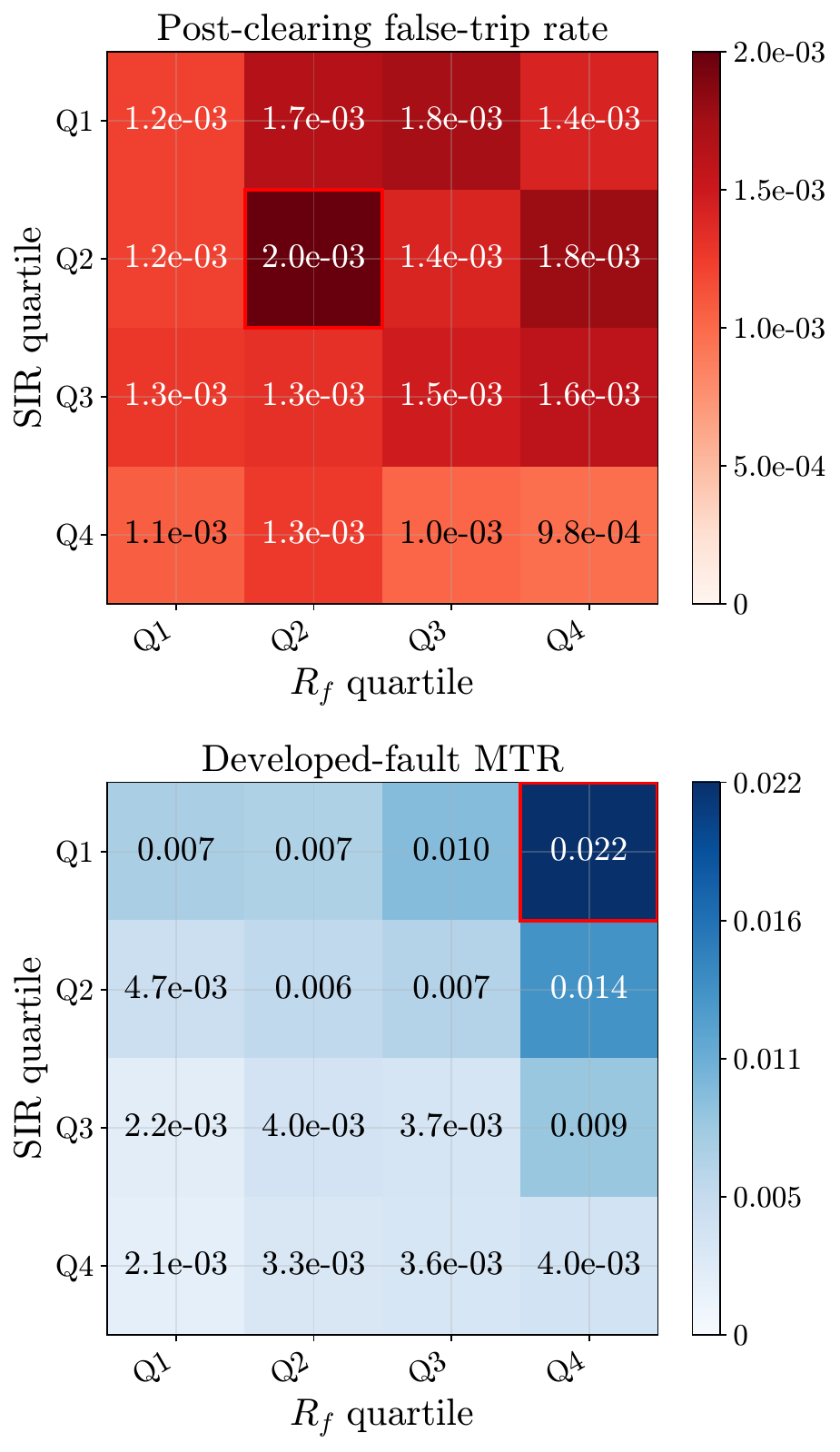}
    \caption{Operating-domain maps of post-clearing false-trip (top) and developed-fault missed-trip (bottom) rates over \acs{sir} and fault-resistance quartiles. Rates are shown as proportions (e.g., $0.010$ denotes $1.0\,\%$) and are averaged over the three \ac{ml} detectors and their seed--fold evaluations within each cell. The developed-fault average is strongly influenced by the larger \acs{logreg} error rates; the per-detector orderings quoted in the text were checked separately. Security is reported post-clearing rather than pooled, because pre-fault decisions precede the existence of the fault resistance. Outlined cells mark the largest visible cell in each panel; with 16 cells per panel this is an inspection aid, not evidence of a worst subgroup.}
    \label{fig:operating-domain}
\end{figure}

The \ac{sir}--fault-resistance maps in Fig.~\ref{fig:operating-domain} give a more consistent picture. For all three \ac{ml} detectors the highest developed-fault \ac{mtr} falls in the highest fault-resistance quartile, and for \ac{cnn} and \ac{logreg} it falls in the lowest-\ac{sir}, highest-$R_f$ cell. Fault resistance is therefore the more consistent of the two descriptors: it orders detector error where measured fault current alone does not. The maps report the post-clearing false-trip rate rather than a pooled security rate because the resistance of the subsequent fault does not physically affect pre-fault measurements. Post-clearing false-trip rates remain below $0.21\,\%$ in every visible cell.

Fault-type stratification separates the detectors more sharply than fault current does. Table~\ref{tab:fault-type} shows that \ac{logreg} misses $4.8\,\%$ of single-line-to-ground and $2.4\,\%$ of line-to-line fault windows, compared with $0.2\,\%$ for double-line-to-ground and $0.5\,\%$ for three-phase faults. \ac{gbt} and \ac{cnn} remain at or below $0.02\,\%$ for every fault type.

\begin{table}[!htbp]
    \centering
    \caption{Missed-trip rate (\%) by fault type on developed-fault windows, mean over three seeds and five outer folds. Sample standard deviations are omitted for width. The small differences between \acs{gbt} and \acs{cnn} are reported descriptively; the table does not establish a statistically reliable ordering. Aggregate variability is reported in Table~\ref{tab:performance}.}
    \label{tab:fault-type}
    \small
    \begin{tabular}{@{}lrrrr@{}}
    \toprule
    \textbf{Model} & \textbf{\acs{slg}} & \textbf{\acs{ll}} & \textbf{\acs{llg}} & \textbf{\acs{lll}} \\
    \midrule
    \acs{gbt}    & $0.009$ & $0.011$ & $0.006$ & $0.004$ \\
    \acs{cnn}    & $0.020$ & $0.012$ & $0.000$ & $0.001$ \\
    \acs{logreg} & $4.83$  & $2.41$  & $0.18$  & $0.52$ \\
    \acs{seqoc}  & $7.36$  & $25.99$ & $20.66$ & $98.57$ \\
    \bottomrule
    \end{tabular}
\end{table}

\Ac{seqoc} exhibits a different error pattern. Constrained to the same $10^{-3}$ security budget it requires a threshold of $0.982\pm0.004$, at which its developed-fault \ac{mtr} is $98.6\,\%$ for balanced three-phase faults against $7.4\,\%$ for single-line-to-ground faults --- an order-of-magnitude separation consistent with zero- and negative-sequence pickup quantities that carry little sustained signal during an ideal balanced fault. Its security--dependability trade-off remains poor across the examined operating points: at $10^{-3}$ it misses $38.3\,\%$ of developed-fault windows and $99.97\,\%$ of onset windows, while its equal-error point reaches a developed-fault \ac{mtr} of $18.2\,\%$ only by operating at a pooled \ac{ftr} of $4.3\,\%$, some 43 times the budget. \Ac{seqoc} therefore serves as a deliberately limited sequence-current reference whose security--dependability trade-off does not intersect the operating region the \ac{ml} detectors occupy, rather than as a complete conventional protection function.

For \ac{cnn} and \ac{logreg}, topology- and phase-specific variations are smaller than those associated with measured fault current and fault type; for \ac{gbt} the topology range is comparable to its fault-type range, and both lie well within the seed--fold spread reported in Table~\ref{tab:performance}. Within the operating conditions represented in PROTECT-90, no single topology state or faulted phase consistently exhibits the highest missed-trip rates across the evaluated \ac{ml} detectors. The complete breakdown is provided in Table~\ref{tab:app-topology-phase}.

\section{Discussion}
\label{sec:discussion}

Nearly identical AUC--ROC values conceal a roughly $280$-fold difference in developed-fault missed-trip rate at matched security, and the separation between detectors increases at fault onset. The developed-fault missed-trip rates of the two nonlinear detectors differ by less than a factor of $1.2$, whereas their onset missed-trip rates differ fivefold. Condition-specific security and dependability therefore reveal differences hidden by pooled classification metrics.

Fault onset is challenging because the first causal window may contain only a partial fault signature, whereas false trips concentrate after clearing, consistent with residual transients. A post-clearing window false-trip rate near $0.1\,\%$ corresponds to at least one spurious assertion in $11$--$19\,\%$ of episodes, showing that window-level security understates episode-level nuisance behaviour. Higher fault resistance within the sampled range is the most consistent operating correlate of missed trips; a monotonic measured-current association appears for only one detector, while topology and phase effects are smaller. \ac{seqoc} instead responds to unbalanced faults but has little sustained response to balanced three-phase faults, making it a targeted reference rather than a complete conventional protection function.

These results support protection-oriented evaluation but not deployment readiness. PROTECT-90 contains simulated faults with idealised, synchronised measurements from all eight relay locations, and security is assessed only before faults and after clearing, not on independent disturbances. Complementary work studies real-world transmission recordings with switching, transformer energisation, and other non-ideal transients~\cite{oelhaf_fault_2026}; related experiments address reduced sampling, missing measurements, and communication outages~\cite{oelhaf_impact_2025,oelhaf_robustness_2026}. Sensor errors, current-transformer saturation, hardware execution, and complete relay logic remain outside the present study. Because the engineered feature set includes sample-to-sample change and high-frequency energy, its behaviour under non-fault switching and energisation transients also requires separate validation.

\section{Conclusion}
\label{sec:conclusion}

This paper developed a protection-oriented assessment of \ac{ml}-based fault detectors beyond aggregate classification performance. Condition-specific security and dependability measures reveal differences that conventional aggregate metrics conceal. Applied to PROTECT-90, the assessment shows that \ac{ml} detector errors are concentrated at fault onset and after fault clearing, that small window-level post-clearing false-trip rates can coexist with episode-level nuisance-assertion rates roughly two orders of magnitude higher, and that the sequence-current reference meets the fixed security budget only at the cost of a 38.3\,\% developed-fault missed-trip rate.

The findings are bounded by the simulated, synchronised, and disturbance-limited setting of the dataset. Broader validation should include independent non-fault disturbances, evolving faults and reclosing, reduced observability, implementation-level timing, additional grids, and complete conventional protection functions. PROTECT-90 and the proposed assessment protocol enable reproducible assessment of whether strong classification performance translates into dependable protection behaviour.

\section*{Acknowledgements}

This project was funded by the Deutsche Forschungsgemeinschaft (DFG, German Research Foundation) -- 535389056.

\section*{Use of generative AI and AI-assisted technologies}

During manuscript preparation, the authors used ChatGPT (OpenAI) and Claude (Anthropic) for drafting support and language refinement. All AI-assisted content was reviewed and edited by the authors, who remain fully responsible for the scientific content and final manuscript.

\section*{References}

\bibliographystyle{IEEEtran}
\bibliography{references}

\clearpage

\appendix

\setcounter{section}{0}
\renewcommand{\thesection}{\Alph{section}}

\setcounter{figure}{0}
\renewcommand{\thefigure}{\thesection.\arabic{figure}}
\setcounter{table}{0}
\renewcommand{\thetable}{\thesection.\arabic{table}}

\section{Additional results}
\label{app:results}

\begin{figure*}[b]
    \centering
    \includegraphics[width=0.9\textwidth]{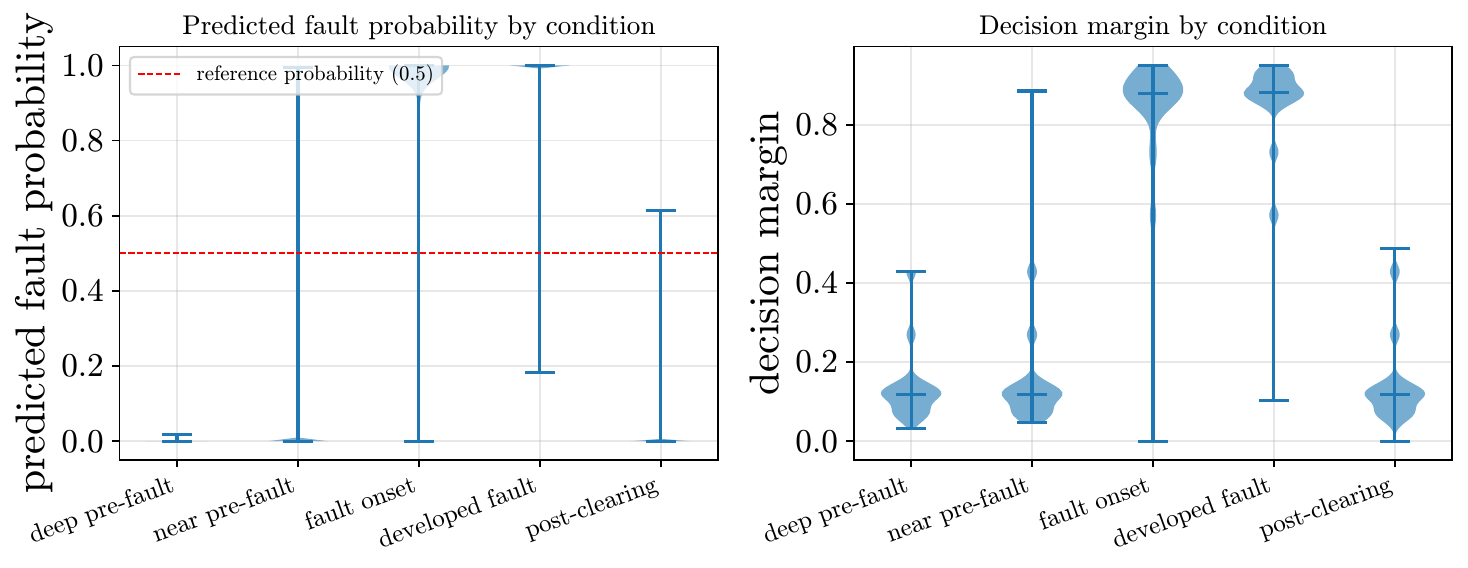}
    \caption{Calibrated fault probability and decision margin by condition for \acs{gbt}. The dashed line marks 0.5 for reference, not the fold-specific operating thresholds. Fault-onset windows show a broad distribution of calibrated probabilities and decision margins.}
    \label{fig:app-margin}
\end{figure*}

Table~\ref{tab:app-security} separates pooled security into deep pre-fault, near pre-fault, and post-clearing conditions. Because the operating points are selected against the same security budget, the detectors differ little in aggregate security; what differs is where the false trips fall. For every detector, false-trip rates are higher after fault clearing than in either pre-fault condition. Values are the mean $\pm$ sample standard deviation over three seeds and five outer folds.

\begin{table}[ht]
    \centering
    \caption{False-trip rate (\%) by non-fault condition.}
    \label{tab:app-security}
    \begin{tabular}{@{}lccc@{}}
    \toprule
    \textbf{Model} & \textbf{Deep pre-fault} & \textbf{Near pre-fault} & \textbf{Post-clearing} \\
    \midrule
    \acs{gbt}    & $0.006\pm0.013$ & $0.011\pm0.023$ & $0.138\pm0.045$ \\
    \acs{cnn}    & $0.023\pm0.042$ & $0.025\pm0.047$ & $0.122\pm0.064$ \\
    \acs{logreg} & $0.002\pm0.009$ & $0.004\pm0.014$ & $0.164\pm0.035$ \\
    \acs{seqoc}  & $0.000\pm0.000$ & $0.000\pm0.000$ & $0.161\pm0.023$ \\
    \bottomrule
    \end{tabular}
\end{table}

Table~\ref{tab:app-topology-phase} reports the missed-trip variation across the eight topology states and three faulted phases. For \ac{cnn} and \ac{logreg} these differences are smaller than those associated with measured fault current and fault type; for \ac{gbt} they are of comparable magnitude. These comparisons are descriptive. The result applies only to the configurations represented in PROTECT-90 and does not establish topology or phase invariance beyond the sampled domain.

\begin{table}[ht]
    \centering
    \caption{Missed-trip rate (\%) by topology and faulted phase; topology gives the range over eight states.}
    \label{tab:app-topology-phase}
    \begin{tabular}{@{}lcc@{}}
    \toprule
    \textbf{Model} & \textbf{Topology range} & \textbf{Faulted phase A/B/C} \\
    \midrule
    \acs{gbt}    & $0.001$--$0.017$ & $0.005/0.008/0.009$ \\
    \acs{cnn}    & $0.005$--$0.012$ & $0.009/0.008/0.007$ \\
    \acs{logreg} & $1.33$--$3.76$ & $1.97/2.05/1.96$ \\
    \bottomrule
    \end{tabular}
\end{table}

\begin{table}[ht]
    \centering
    \caption{Outer-test sensitivity to security budget $\alpha$. ``EE'' denotes the equal-error point. Rates are in percent; post-cl.\ episode rate is the fraction of episodes with at least one post-clearing assertion, and $\tau^\star$ is the mean validation-selected threshold over the 15 seed--fold evaluations.}
    \label{tab:app-sensitivity}
    \footnotesize
    \setlength{\tabcolsep}{3pt}
    \begin{tabular}{@{}lccccc@{}}
    \toprule
    \textbf{Model} & \textbf{$\alpha$} & \textbf{$\tau^\star$} & \textbf{Pooled \acs{ftr}} & \textbf{Dev.\ \acs{mtr}} & \textbf{Post-cl.\ episode rate} \\
    \midrule
    \multirow{4}{*}{\acs{gbt}}
     & $10^{-3}$ & $0.137$ & $0.079$ & $0.007$ & $11.3$ \\
     & $10^{-4}$ & $0.890$ & $0.017$ & $0.195$ & $1.6$ \\
     & $10^{-5}$ & $0.998$ & $0.001$ & $1.575$ & $0.2$ \\
     & EE        & $0.511$ & $0.034$ & $0.031$ & $4.7$ \\
    \midrule
    \multirow{4}{*}{\acs{cnn}}
     & $10^{-3}$ & $0.189$ & $0.078$ & $0.008$ & $12.7$ \\
     & $10^{-4}$ & $0.712$ & $0.009$ & $0.039$ & $2.1$ \\
     & $10^{-5}$ & $0.990$ & $0.001$ & $3.128$ & $0.3$ \\
     & EE        & $0.517$ & $0.018$ & $0.019$ & $4.4$ \\
    \midrule
    \multirow{4}{*}{\acs{logreg}}
     & $10^{-3}$ & $0.871$ & $0.092$ & $1.99$  & $16.8$ \\
     & $10^{-4}$ & $0.997$ & $0.010$ & $16.13$ & $2.2$ \\
     & $10^{-5}$ & $1.000$ & $0.001$ & $45.87$ & $0.3$ \\
     & EE        & $0.277$ & $0.481$ & $0.396$ & $34.6$ \\
    \midrule
    \multirow{4}{*}{\acs{seqoc}}
     & $10^{-3}$ & $0.982$ & $0.089$ & $38.29$ & $19.1$ \\
     & $10^{-4}$ & $0.998$ & $0.007$ & $77.73$ & $1.7$ \\
     & $10^{-5}$ & $1.000$ & $0.001$ & $94.41$ & $0.2$ \\
     & EE        & $0.194$ & $4.343$ & $18.22$ & $99.9$ \\
    \bottomrule
    \end{tabular}
\end{table}

The $10^{-3}$ budget was fixed before outer-test evaluation; the remaining rows are sensitivity analyses and were not used to select the operating point. The inner-validation splits contain 316,106--323,023 security windows (mean 319,592), giving a one-error empirical resolution of approximately $3.1\times10^{-6}$. At $\alpha=10^{-5}$, the selected operating points have zero to three false-positive validation windows. This resolution does not imply precise rare-event estimation with overlapping windows.

\end{document}